\documentclass{article}

\usepackage{arxiv}
\usepackage{multirow}
\usepackage{caption}
\usepackage[utf8]{inputenc} 
\usepackage{hyperref}       
\usepackage{url}            
\usepackage{booktabs}       
\usepackage{amsfonts}       
\usepackage{nicefrac}       
\usepackage{microtype}      
\usepackage{lipsum}
\usepackage{amsmath}
\usepackage{graphicx}
\graphicspath{ {./images/} }
\usepackage{subcaption}
\usepackage{lscape}

\title{DA-DGCEx: Ensuring Validity of Deep Guided Counterfactual Explanations With Distribution-Aware Autoencoder Loss}

\author{
 Jokin Labaien  \\
  Ikerlan Technology Research Centre, \\
Basque Research and Technology Alliance (BRTA), \\
Pº J.M. Arizmediarrieta, 2, 20500 Arrasate/Mondrag\'on, Spain\\
  \texttt{jlabaien@ikerlan.es}
   \And
Xabier de Carlos \\
  Ikerlan Technology Research Centre, \\
Basque Research and Technology Alliance (BRTA), \\
Pº J.M. Arizmediarrieta, 2, 20500 Arrasate/Mondrag\'on, Spain\\
  \texttt{xdecarlos@ikerlan.es} 
  \And
 Ekhi Zugasti \\
  Mondragon University\\
  Data Analysis and Cybersecurity Group, \\
Goiru Kalea, 2, 20500 Arrasate/Mondrag\'on, Spain \\
  \texttt{ezugasti@mondragon.edu}
}

\begin{document}
\maketitle

\begin{abstract}
Deep Learning has become a very valuable tool in different fields, and no one doubts the learning capacity of these models. Nevertheless, since Deep Learning models are often seen as black boxes due to their lack of interpretability, there is a general mistrust in their decision-making process. To find a balance between effectiveness and interpretability, Explainable Artificial Intelligence (XAI) is gaining popularity in recent years, and some of the methods within this area are used to generate counterfactual explanations. The process of generating these explanations generally consists of solving an optimization problem for each input to be explained, which is unfeasible when real-time feedback is needed. To speed up this process, some methods have made use of autoencoders to generate instant counterfactual explanations. Recently, a method called Deep Guided Counterfactual Explanations (DGCEx) has been proposed, which trains an autoencoder attached to a classification model, in order to generate straightforward counterfactual explanations. However, this method does not ensure that the generated counterfactual instances are close to the data manifold, so unrealistic counterfactual instances may be generated. To overcome this issue, this paper presents Distribution Aware Deep Guided Counterfactual Explanations (DA-DGCEx), which adds a term to the DGCEx cost function that penalizes out of distribution counterfactual instances. 
\end{abstract}


\section{Introduction}

In the last decade, with the irruption of Deep Learning (DL), Artificial Intelligence (AI) has risen a step concerning previous years and the evolution of this is being very fast, obtaining great improvements in different areas. Although DL is very effective in many cases, to train these models historical data is needed, which may contain biases. This is a problem because the model can learn the biases inherent in the data, leading to unfair and wrong decisions. Moreover, given that they are complex models with millions of parameters, the decisions taken by them are often difficult to understand because of their lack of interpretability. This is a big drawback in industrial domains, as despite their effectiveness, a lack of confidence in how these models reach conclusions can lead to the use of other, less effective but more interpretable methods. Thus, in recent years an effort has been made to understand these decisions, boosting the research area called \textit{Explainable Artificial Intelligence (XAI)} \cite{arrieta2020explainable}, which aims to find a balance between model effectiveness and model interpretability.

When considering the application of AI systems, it is important to be able to understand the decisions made by these models.  One way to understand these models is through counterfactual explanations. Counterfactual explanations are presented in the form of conditional statements that intend to answer the question ``what would happen if...?". In Machine Learning (ML), a counterfactual explanation consists of describing what should be changed in the input of a model so that the model's prediction changes. A counterfactual explanation can fulfill different properties depending on the context in which it is given \cite{verma2020counterfactual}.

This work is mainly focused on improving some drawbacks that Deep Guided Counterfactual Explanations (DGCEx) has when generating counterfactual explanations. DGCEx is  a simple method that use Autoencoders (AEs) to generate straightforward counterfactual explanations. Among the requirements that Verma et al. \cite{verma2020counterfactual} point out, the counterfactual explanations given with DGCEx satisfy the following conditions:

\begin{itemize}
    \item \textit{Data closeness:} Given an input $\mathbf{x}$, a counterfactual explanation is given as a modified sample $\mathbf{x}_{cf}$. This modified sample, is a minimal perturbation of $\mathbf{x}$, being close to it, i.e.
    \begin{equation}
        \mathbf{x} \approx \mathbf{x}_{cf}.
    \end{equation}
    \item \textit{User-driven:} The counterfactual explanation is user-driven. That is, being $D$ a classification model and $y$ the output assigned by the model to the original input $\mathbf{x}_{cf}$, the user can indicate the desired output $\mathbf{y}_{cf}$, so that the counterfactual explanation is guided to achieve that output, i.e.
    \begin{equation}
        D(\mathbf{x}_{cf}) = \mathbf{y}_{cf}.
    \end{equation}

    \item \textit{Amortized inference:} The counterfactual explanations are straightforward, without solving an optimization problem for each input to be changed. Instead, the explanation model learns to predict the counterfactual. That is, the algorithm can quickly calculate a counterfactual for any given input $\mathbf{x}$. Otherwise, the process of generating explanations is time-consuming.
\end{itemize}

However, we consider that a counterfactual instance, in addition to satisfying the property of data closeness, has to be close to the distribution of the data, i.e. the changes have to be realistic. Data closeness property ensures that the changes are small, but does not ensure that the changes are logical. Therefore, we consider that a counterfactual explanation needs to fulfill the \textit{data manifold closeness} property. In this paper, we propose Distribution Aware Deep Guided Counterfactual Explanations (DA-DGCEx), which includes a term in the loss function of DGCEx that penalizes the changes that falling outside the data manifold.




This article is structured as follows. Section \ref{sec:related_works} describes the related work. Section \ref{sec:background} provides the theoretical background of two baselines methods, DGCEx and Counterfactuals Guided by Prototypes (CFPROTO).  Section \ref{sec:DA-DGCEx} describes Distribution Aware Deep Guided Counterfactual Explanations (DA-DGCEx), the proposed variation of DGCEx. Section \ref{sec:experiments} presents the evaluation metrics, the dataset used, the experimental details and the evaluation of each method. And finally, in Section \ref{sec:conclusions} the conclusions and future works are exposed.

\section{Related Work}
\label{sec:related_works}
In recent years many methods have been proposed regarding the explainability of ML models. Some of these methods use rules to explain other models, which are called rule-based methods, such as Anchors \cite{ribeiro2018anchors} or RuleFit \cite{friedman2008predictive}. In a different way, attribution-based methods attempt to attribute an influence score to each input variable. These methods are the most used ones in the XAI community and these influence attributions can be computed in many ways. One way can be through perturbation-based methods, such as SHAP \cite{lundberg2017unified} or LIME \cite{ribeiro2016should}, which make small variations in the input variables of a model to see how the output varies. Another way can be by gradient-based methods, such as Integrated Gradients \cite{sundararajan2017axiomatic} for example, which compute the attributions as a function of the partial derivatives of the target with respect to the input features.

Additionally, other methods explain model decisions employing counterfactual explanations \cite{stepin2021survey}. Most of these solutions attempt to solve an optimization problem for each input data $\mathbf{x}$ to generate the counterfactual explanation $\mathbf{x}_{cf}$. As a result, the counterfactual explanations are not straightforward, since it is time-consuming to solve the optimization problem for each input $\mathbf{x}$. CEM \cite{dhurandhar2018explanations} was one of the first methods regarding counterfactual explanations. In CEM, the explanations are minimal changes that have to be made to an input so that the model prediction changes to its nearest class. Similarly, in  \cite{van2019interpretable}, the authors introduce a term regarding class prototypes, which speeds up the explanation process and also allows to specify the class to which the input has to be changed. However, in this case, the process also consists of solving an optimization problem, so the explanations are not straightforward. Recently, DiCE \cite{mothilal2020explaining}  has been proposed, which instead of generating a single counterfactual explanation for each input, it generates a set of $k$ different counterfactual explanations. Nonetheless, DiCE is still limited to binary classification problems and the explanations are not straightforward.

Since these methods are time-consuming when generating explanations, are hardly applicable in scenarios where a quick response is needed. Therefore, recently, emphasis is being placed on \textit{amortized inference methods}, which consist of models trained to generate counterfactual explanations instantaneously. This models are usually generative models, such as \textit{Generative Adversarial Networks (GANs)} \cite{nemirovsky2020countergan,liu2019generative} or \textit{Variational Autoencoders (VAEs)} \cite{mahajan2019preserving}. Nevertheless, generative models have their benefits and drawbacks.  On the one hand, VAEs are easier to train compared to GANs, but the latent space is restricted to follow a prior distribution and the reconstructions are usually of lower quality than GANs. GANs, however, are often difficult to train as both the generator model and the discriminator model are trained simultaneously in a game and the improvements to one model come at the expense of the other model. Recently it has been proposed to use a simple AE to generate counterfactual explanations. Balabsubramanian et al. \cite{balasubramanian2020latent} propose a simple method that considers binary classification problems. In this method, the counterfactual explanations are obtained by minimally changing the latent representation of the encoder so that the prediction given by a classifier $D$ to the decoded sample reaches a predefined target probability $p$. DGCEx is an amortized inference method that uses AEs for generating straightforward counterfactual explanations, that can be applied in multiclass scenarios. 

In this work, DA-DGCEx is proposed, which includes a term in the loss function of DGCEx that penalizes illogical changes in the input, penalizing instances falling outside the data distribution. For the experiments has been considered the MNIST handwritten digits classification problem. In this use case CFPROTO, DGCEx, and DA-DGCEx have been evaluated taking into account the metrics IM1 and IM2 proposed in \cite{van2019interpretable} and the speed at which the counterfactual instances are generated. 

\section{Background} \label{sec:background}

Next, the the theoretical background of Counterfactuals Guided by Prototypes and Deep Guided Counterfactual Explanations is provided. 

\subsection{Counterfactuals Guided by Prototypes}

Counterfactuals Guided by Prototypes (CFPROTO) \cite{van2019interpretable} is a model-agnostic method to find interpretable counterfactual explanations for classifier predictions by using class prototypes.

Let $\mathbf{x}$ be an input of a black-box model (to avoid notational inconsistencies, this model is referred as the discriminator (D)), $D(\mathbf{x})$ the prediction given and $\mathbf{y}$ its corresponding class. Let $\mathrm{AE}(\cdot)$ be an AE trained for reconstructing an input. Denoting $\mathcal{X}/\mathbf{x}$ to the space of missing parts with respect to $\mathbf{x}$, CFPROTO attempts finding a counterfactual instance $\mathbf{x}_{cf} = \mathbf{x}+\delta$, such that the predicted class changes, i.e. $\arg \max _{i}\left[D\left(\mathbf{x}\right)\right]_{i} \neq \arg \max _{i}\left[D\left(\mathbf{x}_{cf} \right)\right]_{i}$. Finding these perturbations implies optimizing the following objective function 

\begin{equation}
\min _{\delta} c \cdot f_{\kappa}\left(\mathbf{x}_{cf}\right)+f_{\text {dist }}(\delta) + f_{\text{AE}}\left(\mathbf{x}_{cf}\right) + f_{\text{proto}}(\mathbf{x}_{cf}).
\end{equation}

The function $f_{\kappa}$  encourages the predicted class of $\mathbf{x}_{cf}$ to be different to the original class $\mathbf{y}$. Dhurandhar et al. \cite{dhurandhar2018explanations} define this function as

\begin{equation}
\begin{aligned}
L_{\text {pred }} &:=f_{\kappa}\left(\mathbf{x}_{cf}\right) \\
&=\max (\left[D(\mathbf{x}_{cf})\right]_{y}-\max _{i \neq y}\left[D(\mathbf{x}_{cf})\right]_{i},-\kappa)
\end{aligned}
\end{equation}

where $\left[D(\mathbf{x}_{cf})\right]_i$ is the $i$-th class prediction probability and $\kappa \geq 0$ controls the separation between  $\left[D(\mathbf{x}_{cf})\right]_i$ and  $\left[D(\mathbf{x}_{cf})\right]_{\mathbf{y}}$. Moreover, the term $f_{\text{dist}}(\delta)$  is an elastic net regularizer, which minimizes the distance between $\mathbf{x}$ and  $\mathbf{x}_{cf}$, ensuring sparsity, i.e.

\begin{equation}
f_{\text {dist }}(\delta)=\beta \cdot\|\delta\|_{1}+\|\delta\|_{2}^{2}=\beta \cdot L_{1}+L_{2}.
\end{equation}

Moreover, the term $f_{\text{AE}}\left(\mathbf{x}, \delta\right)$ penalizes out-of-distribution counterfactual instances, by minimizing the distance between the counterfactual $\mathbf{x}_{cf}$ and its reconstruction using an AE fitted with the training set, i.e.
\begin{equation}
\begin{aligned}
    L_{\mathrm{AE}}&:= f_{\text{AE}}(\mathbf{x}_{cf})\\
    &=\gamma \cdot\left\|\mathbf{x}_{cf}-\mathrm{AE}\left(\mathbf{x}_{cf}\right)\right\|_{2}^{2}
\end{aligned}
\end{equation}

Finally, to speed up the counterfactual search process and to guide the perturbations towards a predefined class, Van Looveren et al. \cite{van2019interpretable} include a prototype loss term $f_{\text{proto}}$ . Let $\operatorname{ENC}(\cdot)$ be the encoder of the AE. Then, for each class $i$, the instances belonging to that class are encoded using $\operatorname{ENC}$ and they are ordered by increasing $L_2$ distance to $\operatorname{ENC}(\mathbf{x})$. After that, the class prototype is defined as the average encoding over the $K$ nearest instances in the
latent space with the same class label:
\begin{equation}
\operatorname{proto}_{i}:=\frac{1}{K} \sum_{k=1}^{K} \operatorname{ENC}\left(\mathbf{x}_{k}^{i}\right)
\end{equation}
for the ordered $\left\{\mathbf{x}_{k}^{i}\right\}_{k=1}^{K}$ in class $i$. Then, the prototype loss can be defined as

\begin{equation}
\begin{aligned}
    L_{\mathrm{proto}}&:= f_{\text{proto}}(\mathbf{x}_{cf})\\
    &=\theta \cdot\left\|\operatorname{ENC}(\mathbf{x}_{cf})-\operatorname{proto}_j\right\|_{2}^{2}.
\end{aligned}
\end{equation}
This term guides the perturbations towards the prototype $\operatorname{proto}_{\mathbf{y}_{cf} \neq \mathbf{y}}$, speeding up the
counterfactual search process towards the average encoding
of class $\mathbf{y}_{cf}$, which can be the class of the nearest prototype or a class predefined by the user. Therefore, CFPROTO aims to minimize the following objective function:
\begin{equation}
L=c \cdot L_{\mathrm{pred}}+\beta \cdot L_{1}+L_{2}+L_{\mathrm{AE}}+L_{\text {proto }}
\end{equation}
\subsection{Deep Guided Counterfactual Explanations} \label{sec:DGCEx}

\textbf{D}eep \textbf{G}uided \textbf{C}ounterfactual \textbf{Ex}planations (\textbf{DGCEx}) \cite{labaien2021dadgcex} is a deep learning based algorithm that uses an AE to  provide counterfactual explanations of what changes should be made in an input $\mathbf{x}$ of a classification model to switch its class to another predefined class $\mathbf{y}_{cf}$.

\begin{figure}[!ht]
    \centering
    \includegraphics[width=0.7\textwidth]{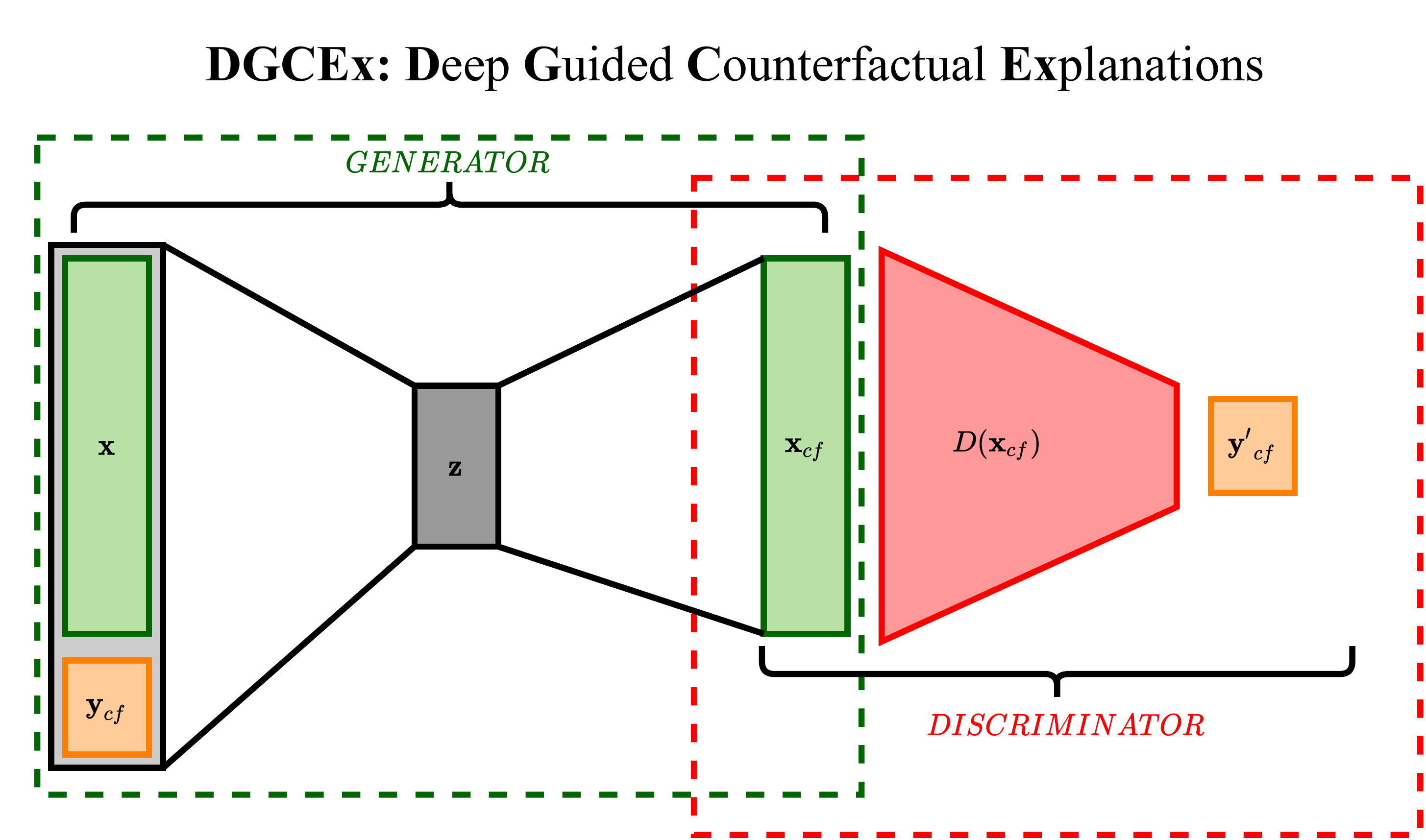}
    \caption{ Deep Guided Counterfactual Explanations.}
    \label{fig:DGCEx}
\end{figure}

As shown in Figure \ref{fig:DGCEx}, DGCEx is a simple algorithm that is composed of two deep learning models: the \textit{generator} (G), which is an AE used to generate counterfactual explanations, and the \textit{discriminator} (D), which is a  model trained to classify the input between different classes. The essence of DGCEx is how G is trained. 

Let $\mathcal{D}=\{(\mathbf{x}^{(i)},\mathbf{y}^{(i)})\}_{i=1}^{N}$ be a dataset of inputs paired with their corresponding label. Then, G is trained with the objective of making minimal changes to an input $\mathbf{x}$, so that the class of the changed sample $\mathbf{x}_{cf}$ is equal to the counterfactual class $\mathbf{y}_{cf}$ defined by the user. That is, given an input $\mathbf{x}\in \mathcal{D}$ and a counterfactual class $\mathbf{y}_{cf}$,
    \begin{equation}\label{eq:dgcex_conditions}
        G(\mathbf{x}) \approx \mathbf{x} \quad \text{and} \quad D(G(\mathbf{x}))=\mathbf{y}_{cf}
    \end{equation}
    For this purpose, only the weights corresponding to G are trained but making use of the capacities that D has to classify the inputs. That is, when training G, the weights of D are frozen and G is trained to minimize the following cost function
    
    \begin{equation}\label{eq:dgcex_loss}
        \mathcal{L}^{\text{G,D}}(\mathbf{x},\mathbf{x}_{cf},\mathbf{y}_{cf},\mathbf{y'}_{cf}) =  \alpha \cdot \mathcal{L}^{\text{G}}(\mathbf{x},\mathbf{x}_{cf}) + \beta \cdot \mathcal{L}^{\text{D}}(\mathbf{y}_{cf},\mathbf{y'}_{cf})
    \end{equation}
    where $\alpha, \beta > 0$ are regularization coefficients, $\mathcal{L}^{G}(\mathbf{x},\mathbf{x}_{cf})$  measures the distance between $\mathbf{x}$ and the counterfactual sample $\mathbf{x}_{cf}$ , which ensures that the changes are minimal, and $\mathcal{L}^{D}(\mathbf{y}_{cf},\mathbf{y'}_{cf})$ measures the distance between the predefined counterfactual class $\mathbf{y}_{cf}$ and the prediction given by D to $\mathbf{x}_{cf}$, i.e. $\mathbf{y}'_{cf}$, which ensures that the input $\mathbf{x}$ is changed to the class $\mathbf{y}_{cf}$ defined by the user. For G to be able to change every input $\mathbf{x} \in \mathcal{D}$ to any specified class $\mathbf{y}_{cf}\in \mathcal{Y}$, being $\mathcal{Y}$ the family of any possible class that $\mathbf{x}$ could belong to, while training G, each sample $\mathbf{x} \in \mathcal{D}$ is paired with any possible $\mathbf{y}_{cf} \in \mathcal{Y}$, such that $\mathbf{y}_{cf} \neq \mathbf{y}$, being $\mathbf{y}$ the original class. 
    
    

\section{Distribution Aware DGCEX} \label{sec:DA-DGCEx}
As pointed out before, the \textit{data closeness} property does not imply the \textit{data manifold closeness} property, since the changes made to an input can be minimal, ensuring data closeness, but not realistic. Therefore, Distribution Aware Deep Guided Counterfactual Explanations (DA-DGCEx) overcomes this problem by adding a term to the loss function that penalizes unrealistic changes. 

\begin{figure}[!ht]
    \centering
    \includegraphics[width=0.7\textwidth]{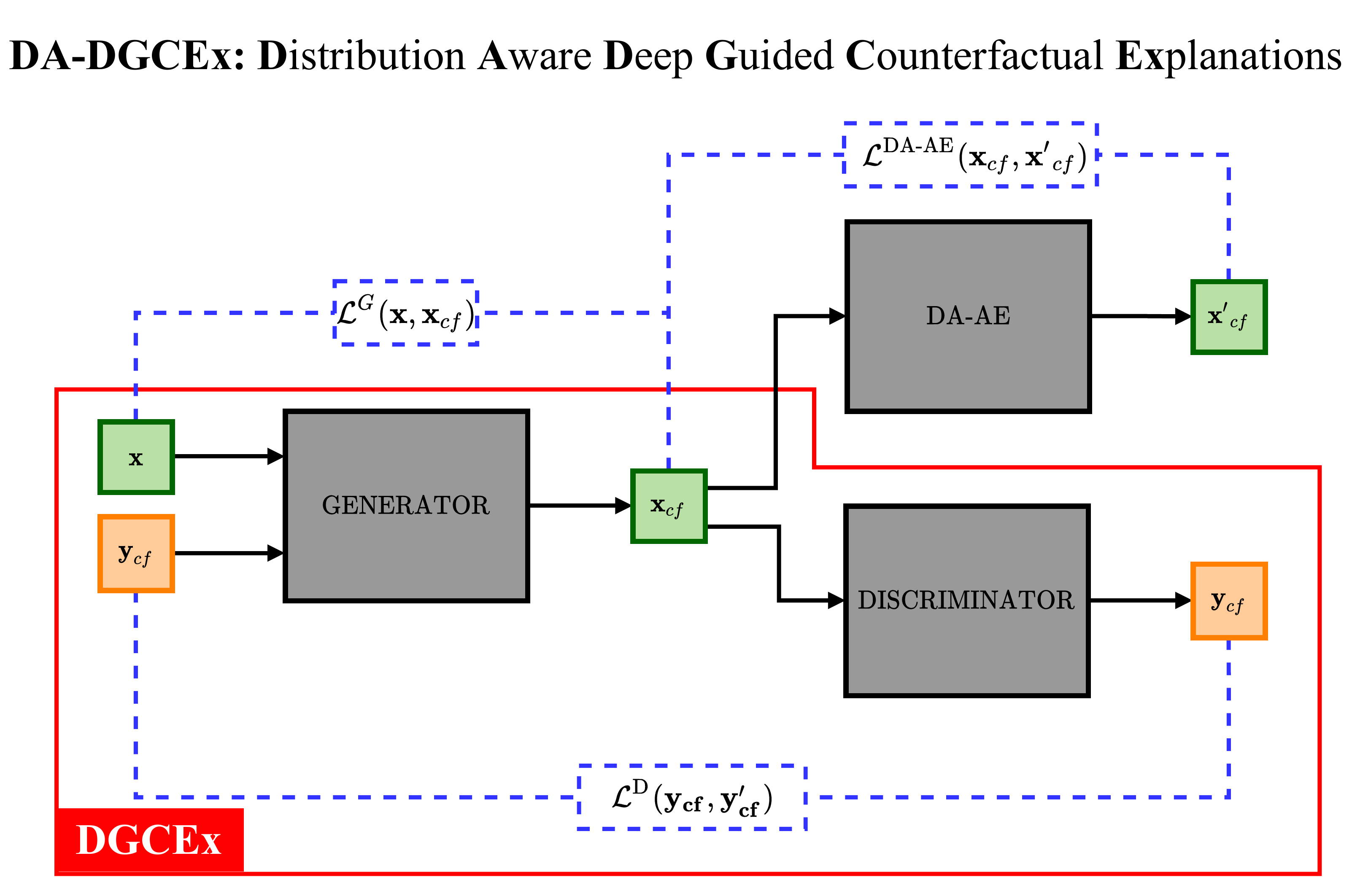}
    \caption{ Distribution Aware Deep Guided Counterfactual Explanations.}
    \label{fig:DADGCEx}
\end{figure}

DA-DGCEx is similar to DGCEx but an AE is added to penalize out-of-distribution counterfactual instances. In this method, the output of the generator, besides being the input of the discriminator, is also the input of an AE that is trained to reconstruct instances of all possible classes. This AE is called Distribution Aware Autoencoder (DA-AE).  For training G, both the discriminator (D) and DA-AE are frozen and the training is done in the same way as in DGCEx, i.e. pairing every input $\mathbf{x}$ with all possible classes $\mathbf{y}_{cf}$ to which $\mathbf{x}$ can be modified. In this case, an extra term is added to the cost function regarding DA-AE, and the optimization problem to minimize becomes

\begin{equation}\label{eq:dadgcex_loss}
    \mathcal{L}^{\text{G,D,DA-AE}}(\mathbf{x},\mathbf{x}_{cf},\mathbf{x'}_{cf}\mathbf{y}_{cf},\mathbf{y'}_{cf}) =  \alpha \cdot \mathcal{L}^{\text{G}}(\mathbf{x},\mathbf{x}_{cf}) + \beta \cdot \mathcal{L}^{\text{D}}(\mathbf{y}_{cf},\mathbf{y'}_{cf}) + \gamma \cdot \mathcal{L}^{\text{DA-AE}}(\mathbf{x}_{cf},\mathbf{x'}_{cf})
\end{equation}

where $\alpha, \beta, \gamma > 0 $ are the regularization coefficients. The term $\mathcal{L}^{\text{DA-AE}}(\mathbf{x}_{cf},\mathbf{x'}_{cf})$ is the reconstruction error of the counterfactual instance $\mathbf{x}_{cf}$ when using DA-AE. Since DA-AE has been trained with instances of the dataset $\mathcal{D}$, the reconstruction error will be smaller when $\mathbf{x}_{cf}$ belongs to the data manifold, so $\mathcal{L}^{\text{DA-AE}}(\mathbf{x}_{cf},\mathbf{x'}_{cf})$ penalizes unrealistic changes of the input $\mathbf{x}$, ensuring the \textit{data manifold closeness} property. 

\section{Experiments} \label{sec:experiments}
This section describes the dataset used in the experiments, the evaluation metrics, the model architectures, the results obtained, and the experimental framework.

\subsection{MNIST: Handwritten Digits}
MNIST \cite{lecun2010mnist} is a widely used handwritten digit classification public dataset. This dataset contains 70.000 small 28×28 pixel square grayscale images of simple handwritten digits between 0 and 9. In the conducted experiments, 60.000 images have been used to train the models and the remaining 10.000 for evaluation.
\subsection{Evaluation}
To assess the interpretability of the counterfactuals generated with each method, two metrics proposed by Van Looveren et al. \cite{van2019interpretable}, IM1 and IM2, have been used. On the one hand, IM1 measures the ratio between the reconstruction error of the counterfactual $\mathbf{x}_{cf}$ using an AE specifically trained with instances of the counterfactual class ($\mathrm{AE}_{\mathbf{y}_{cf}}$) and another AE trained with instances of the original class ($\mathrm{AE}_{\mathbf{y}}$). 

\begin{equation}
\operatorname{IM1} \left(\mathrm{AE}_{\mathbf{y}_{cf}}, \mathrm{AE}_{\mathbf{y}}, \mathbf{x}_{cf}\right):=\frac{\left\|\mathbf{x}_{cf}-\mathrm{AE}_{\mathbf{y}_{cf}}\left(\mathbf{x}_{cf}\right)\right\|_{2}^{2}}{\left\|\mathbf{x}_{cf}-\mathrm{AE}_{\mathbf{y}}\left(\mathbf{x}_{cf}\right)\right\|_{2}^{2}+\epsilon}
\end{equation}

A smaller value of IM1 means that the counterfactual instance is easier to reconstruct with the AE trained only with instances of the class $\mathbf{y}_{cf}$ than with the AE trained only with instances of the original class $t_0$. This means that the counterfactual instance is closer to the data manifold of the $\mathbf{y}_{cf}$ class instances than to the ones belonging to the original class  $t_0$. On the other hand, IM2 measures how close the counterfactual instance is to the data manifold. To do so, it compares the reconstructions of the counterfactual instances when using an AE trained with instances of all classes (AE) and the AE trained with instances of the counterfactual class ($\mathrm{AE}_{\mathbf{y}_{cf}}$). To make the metric comparable across all classes, IM2 is scaled with the $L_1$ norm of $\mathbf{x}_{cf}$.

\begin{equation}
\operatorname{IM2} \left(\mathrm{AE}_{\mathbf{y}_{cf}}, \mathrm{AE}, \mathbf{x}_{\mathrm{cf}}\right):=\frac{\left\|\mathrm{AE}_{\mathbf{y}_{cf}}\left(\mathbf{x}_{cf}\right)-\mathrm{AE}\left(\mathbf{x}_{cf}\right)\right\|_{2}^{2}}{\left\|\mathbf{x}_{cf}\right\|_{1}+\epsilon}
\end{equation}
When the reconstruction of $\mathbf{x}_{cf}$ by AE and by $\mathrm{AE}_{\mathbf{y}_{cf}}$ is similar, IM2 is low. A lower value of IM2 makes the counterfactual instance more interpretable since the data distribution of the counterfactual class $\mathbf{y}_{cf}$ describes $\mathbf{x}_{cf}$ as good as the distribution over all classes.


\subsection{Model Architectures}
This section briefly explains the architectures used for the discriminator, the autoencoder, and the generator. 
\paragraph{\textit{Discriminator}.} The discriminator used has the same architecture as the one used in \cite{van2019interpretable}. That is, the model consists of 2 convolutional layers with 32 and 64 2x2 filters respectively, with ReLU activation function. Besides, a 2x2 MaxPooling layer is applied after each convolutional layer to reduce dimensionality and avoid overfitting. Dropout with fraction 30\% is applied
during training. After the second MaxPooling layer, the output is flattened to apply a fully connected layer with 256 units, ReLU activation function, and 50\% dropout. Finally, another fully connected layer with 10 units and a softmax activation function is used for classification. The model is trained with an Adam optimizer for 10 epochs with batch-size 32, reaching a test accuracy of 99.1\%.

\paragraph{\textit{Autoencoder}.} The encoder is composed of two convolutional layers with 32 and 64 2x2 filters respectively, with ReLU activation function. The output of the second convolutional layer is flattened to feed a fully connected layer with 16 units, from which a 16-dimensional latent vector $\mathbf{z}$ is obtained. Then, this vector is feed to a fully connected layer which transforms the  $\mathbf{z}$ vector into a vector of the same dimensionality as the flattened output of the second convolutional layer. This vector is reshaped into a 7x7x64 tensor to initialize the transposed convolutions. The decoder has three transpose convolution layers, with 64, 32, and 1 2x2 filters respectively. The first two transpose convolution layers have a ReLU activation function and the last one has a sigmoid.

\paragraph{\textit{Generator}.} The architecture of the generator is almost the same as the AE, except that in this case the latent vector $\mathbf{z}$ is concatenated with the one-hot representation of the class to which the input has to be changed, i.e. $\mathbf{y}_{cf}$.

Regarding the value of the regularizers, in CFPROTO we have used those recommended by Van Looveren et al. \cite{van2019interpretable}, that is: $\beta = 0.1, \gamma = 100, \theta = 100$ and $c = 1$. Note that $c$ is decreased when a solution is found, aiming to find a perturbation closer to the original instance, and it is increased when the current solution is not valid. In DGCEx, the  $\beta$ parameter has been set to 1 and the $\alpha$ has been set to 10. Finally, in DA-DGCEx, the $\alpha$ and $\beta$ chosen are equal to the ones in DGCEx and $\gamma = 10$.

In the conducted experiments, the following losses are used to train DA-DGCEx: the term $\mathcal{L}^G$ is the $L_2$ norm between $\mathbf{x}$ and $\mathbf{x}_{cf}$, the term $\mathcal{L}^D$ is the \textit{Categorical Cross-Entropy (CCF)}
between $\mathbf{y}$ and $\mathbf{y}_{cf}$, and the last term, i.e. $\mathcal{L}^{\text{DA-AE}}$, measures the $L_2$ norm between $\mathbf{x}_{cf}$ and $\mathbf{x'}_{cf}$.
\subsection{Results}

This section presents the results obtained in the MNIST use case with each of the methods. To evaluate the interpretability, the IM1 and IM2 metrics explained in the previous section have been used. Moreover, the time taken by each method to generate the counterfactual instances has been calculated. The variability of the algorithms has also been taken into account, and which has been represented with the \textit{Interquartile Range (IQR)}. The IQR is defined as the difference between the first quartile ($\mathrm{Q}_1$) and the third quartile ($\mathrm{Q}_2$), i.e., 

\begin{equation}
\mathrm{IQR}=\mathrm{Q}_{3}-\mathrm{Q}_{1}
\end{equation}

\begin{figure}[!ht]
    \centering
    \includegraphics[width=0.6\textwidth]{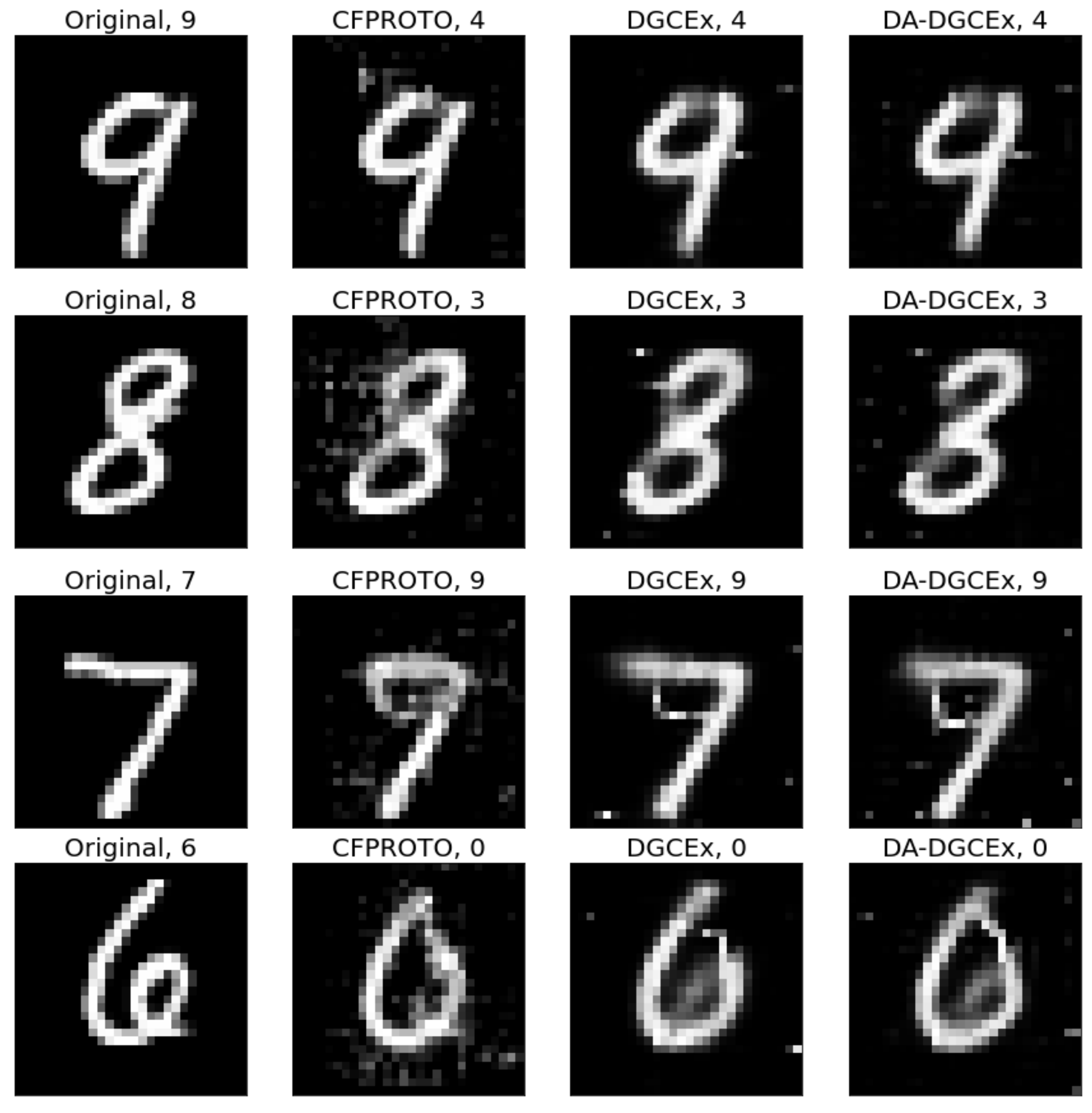}
    \captionsetup{width=.6\textwidth,skip=5pt}

    \caption{Examples of original and counterfactual instances generated with CFPROTO, DGCEx and DA-DGCEx.}
    \label{fig:comparison}
\end{figure}

In Figure \ref{fig:comparison} we have selected some examples of the results obtained with each method. In the first column of the image are placed the original instances with their respective labels, and in the second, third, and fourth columns are placed the counterfactual instances generated with CFPROTO, DGCEx, and DA-DGCEx, respectively. The first row shows how the original instance representing the digit 9 is converted into a 4 by each of the algorithms. In this example, CFPROTO modifies a small part of the upper side of the 9 by deleting a few pixels that make the 9 become a 4. In DGCEx and DA-DGCEx these changes are somewhat larger, but the result looks more like a 4 compared to CFPROTO. In the second one, an 8 is converted into a 3. In this case, also something similar happens, the counterfactual instance generated by DGCEx and especially by DA-DGCEx is clearer than the one generated by CFPROTO. In the following example, a 7 is converted to a 9. In this case, all methods produce counterfactual instances quite similar to a 9, although perhaps the 9 generated by CFPROTO is more realistic. Finally, a 6 is changed to a 0 by deleting the pixels of the lower circle of the 6, and adding new pixels for joining with the upper part, creating the circle. 

To evaluate the models, a counterfactual instance has been generated for each test sample. For a fair evaluation, CFRPOTO has been used first to generate instances of the nearest prototype class $\mathbf{y}_{cf}$ and then DGCEx and DA-DGCEx have been used to generate instances of the same class. The results obtained with each method are summarized in Table \ref{tab:numerical_comparison}, which shows the mean and IQR of each method in terms of IM1, IM2, and speed. As for IM1, CFPROTO and DA-DGCEx outperform DGCEx. The results of CFPROTO and DA-DGCEx in terms of this metric are similar, but CFPROTO has less variability, as the IQR is 0.39, while the IQR of DA-DGCEx is 0.58. As for IM2, DA-DGCEx outperforms the other two algorithms. Comparing CFPROTO with DGCEx it can be seen that the mean of the values obtained with DGCEx is 0.08 points lower than those obtained with CFPROTO and the variability is also lower. The DGCEx results are significantly improved by introducing the term that penalizes the out-of-distribution changes, and the mean IM2 values decrease from 1.12 in DGCEx to 0.91 in DA-DGCEx, also decreasing the variability.  Moreover, when analyzing the time taken by each algorithm to generate the counterfactual instances, there is no comparison between CFPROTO and the other two algorithms, since while CFPROTO takes 17.40 seconds per iteration with a variability of 0.09 seconds, DGCEx and DA-DGCEx take 0.14 seconds with a variability of 0.01 seconds. Note that DGCEx and DA-DGCEx take the same time since at inference time has the same architecture. 



\begin{table}[ht!]
\centering
\resizebox{0.65\textwidth}{!}{%
\begin{tabular}{l|ll|ll|ll}
\toprule
\multicolumn{1}{c|}{\multirow{2}{*}{\textbf{Method}}} & \multicolumn{2}{c|}{\textbf{IM1}} & \multicolumn{2}{c|}{\textbf{IM2 (x10)}}  & \multicolumn{2}{c}{\textbf{Speed (s/it)}} \\
\multicolumn{1}{c|}{}                        & Mean        & IQR        & \multicolumn{1}{c}{Mean} & IQR  & \multicolumn{1}{c}{Mean}  & IQR  \\ \midrule
\textit{CFPROTO}                             &\textbf{1.20}       & \textbf{0.39}       & 1.20                     & 0.83 & 17.40                     & 0.09 \\
\textit{DGCEx}                               & 1.40        & 0.63       & 1.12                     & 0.74 & \textbf{0.14}                      & \textbf{0.01} \\
\textit{DA-DGCEx}                            & 1.21        & 0.58       & \textbf{0.91}                     & \textbf{0.63} & \textbf{0.14}                      & \textbf{0.01} \\ \bottomrule
\end{tabular}%
}
\captionsetup{width=.65\textwidth,skip=5pt}

\caption{Summary of the results obtained in test for each
method.}
\label{tab:numerical_comparison}
\end{table}

\begin{figure}[ht!]
     \centering
     \begin{subfigure}[b]{0.45\textwidth}
         \centering
         \includegraphics[width = 7cm,height=5cm]{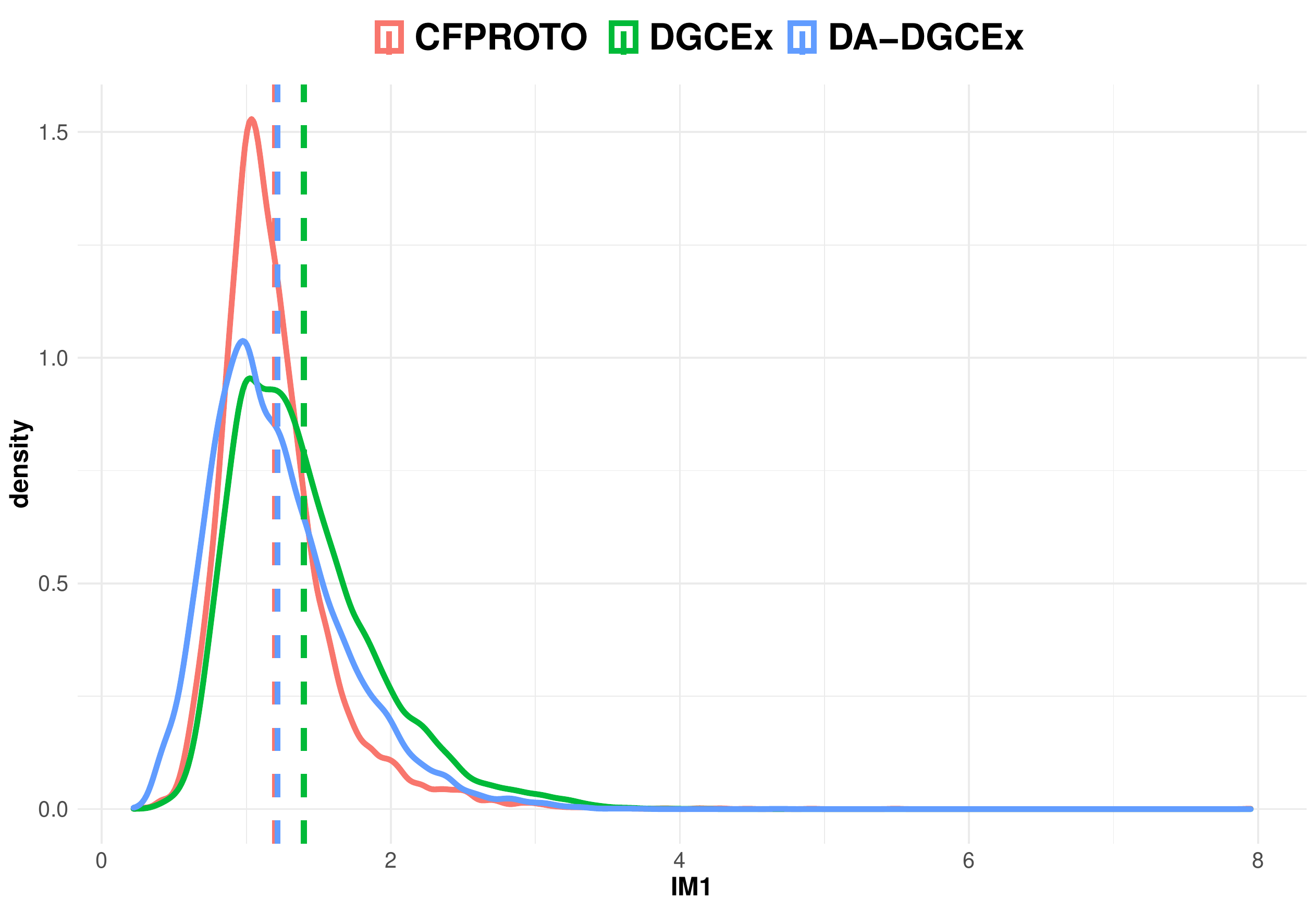}
         \caption{IM1 distribution plots.}
         \label{fig:IM1_dist}
     \end{subfigure}
     \hfill
     \begin{subfigure}[b]{0.45\textwidth}
         \centering
         \includegraphics[width = 7cm,height=5cm]{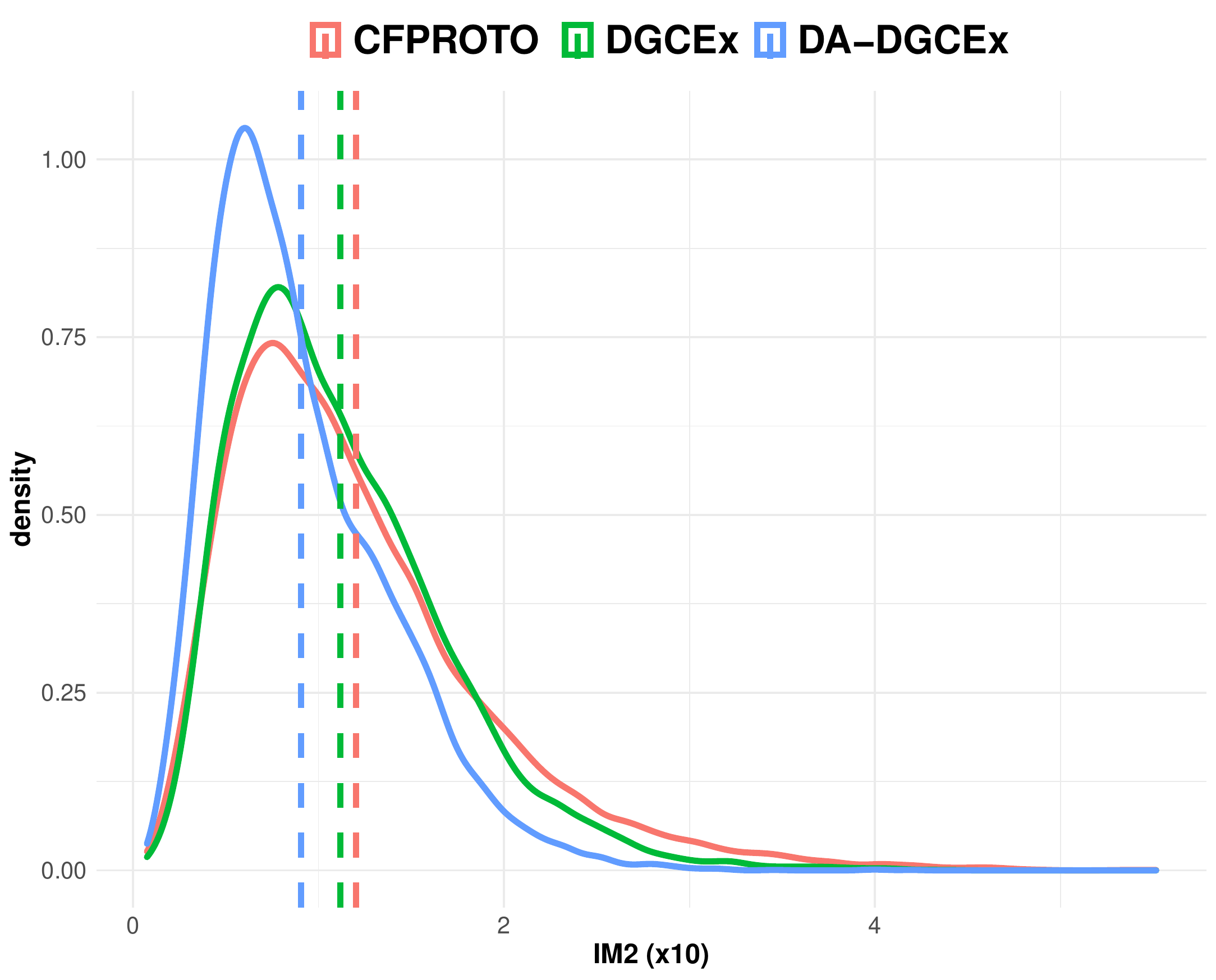}
         \caption{IM2 (x10) distribution plots.}
         \label{fig:IM2_dist}
     \end{subfigure}

        \caption{Distribution plots for each method in terms of IM1 and IM2 (x10) metrics. The dashed lines represent the mean.}
        \label{fig:distribution_plots}
\end{figure}

For the comparison of models, using statistical tests on very large samples does not make much sense, as p-values quickly go to 0, which may lead the researcher to claim support for results of no practical significance \cite{lin2013research}.  Since we have 10.000 instances in the test set, the distribution of the IM1 and IM2 values obtained will be used to make the model comparisons. In Figure \ref{fig:distribution_plots} the distributions of each method in terms of IM1 and IM2 have been shown.  These distributions support the conclusions drawn above from the Table \ref{tab:numerical_comparison}. That is, in terms of IM1 CFPROTO and DA-DGCEx are similar concerning the mean (which is represented with dashed vertical lines) and DGCEx and DA-DGCEx have more variability. And as for IM2, the mean and the variability when using DA-DGCEx is lower compared to the other two methods. From these distributions interesting numerical conclusions can be drawn regarding the comparison of the models, such as the probability that a counterfactual instance gives lower values in terms of IM1 or IM2 when it is generated with one method is lower than using another method. These comparisons are shown in Table \ref{tab:prob_comparison}. In this table, for example, the 0.66 in the first row and second column represents that the probability that the IM1 value of an instance generated by CFPROTO is less than an instance generated by DGCEx is 0.66. This shows that DA-DGCEx improves drastically over DGCEx since an instance generated by DA-DGCEx has a probability of having lower IM1 and IM2 of about 0.75 compared to the one generated with DGCEx. Comparing DA-DGCEx with CFPROTO, in terms of IM2, the probability of obtaining better counterfactual instances with DA-DGCEx is 0.62 with respect to CFPROTO. As for IM1, CFPROTO and DA-DGCEx have the same probability that the value will be lower using one method or the other. 

\begin{table}[ht!]
\centering
\resizebox{0.7\textwidth}{!}{%
\begin{tabular}{l|ll|ll|ll}
\toprule
\multicolumn{1}{c|}{\multirow{2}{*}{\textbf{Method}}} & \multicolumn{2}{l|}{\textit{CFPROTO}} & \multicolumn{2}{c|}{\textit{DGCEx}} & \multicolumn{2}{c}{\textit{DA-DGCEx}} \\
\multicolumn{1}{c|}{}                                 & IM1               & IM2               & IM1              & IM2              & IM1               & IM2               \\ \midrule
\textit{CFPROTO}                                      & \multicolumn{2}{c|}{-}                & 0.66             & 0.53             & 0.50              & 0.38              \\
\textit{DGCEx}                                        & 0.34              & 0.47              & \multicolumn{2}{c|}{-}              & 0.26              & 0.24              \\
\textit{DA-DGCEx}                                     & 0.50              & 0.62              & 0.74             & 0.76             & \multicolumn{2}{c}{-}                 \\ \bottomrule
\end{tabular} }%
\captionsetup{width=.7\textwidth,skip=5pt}

\caption{This table represents the probabilities that a counterfactual instance has a lower IM1 or IM2 depending on the method used.
}
\label{tab:prob_comparison}
\end{table}

\begin{figure}[!ht]
    \centering
    \includegraphics[width=0.5\textwidth]{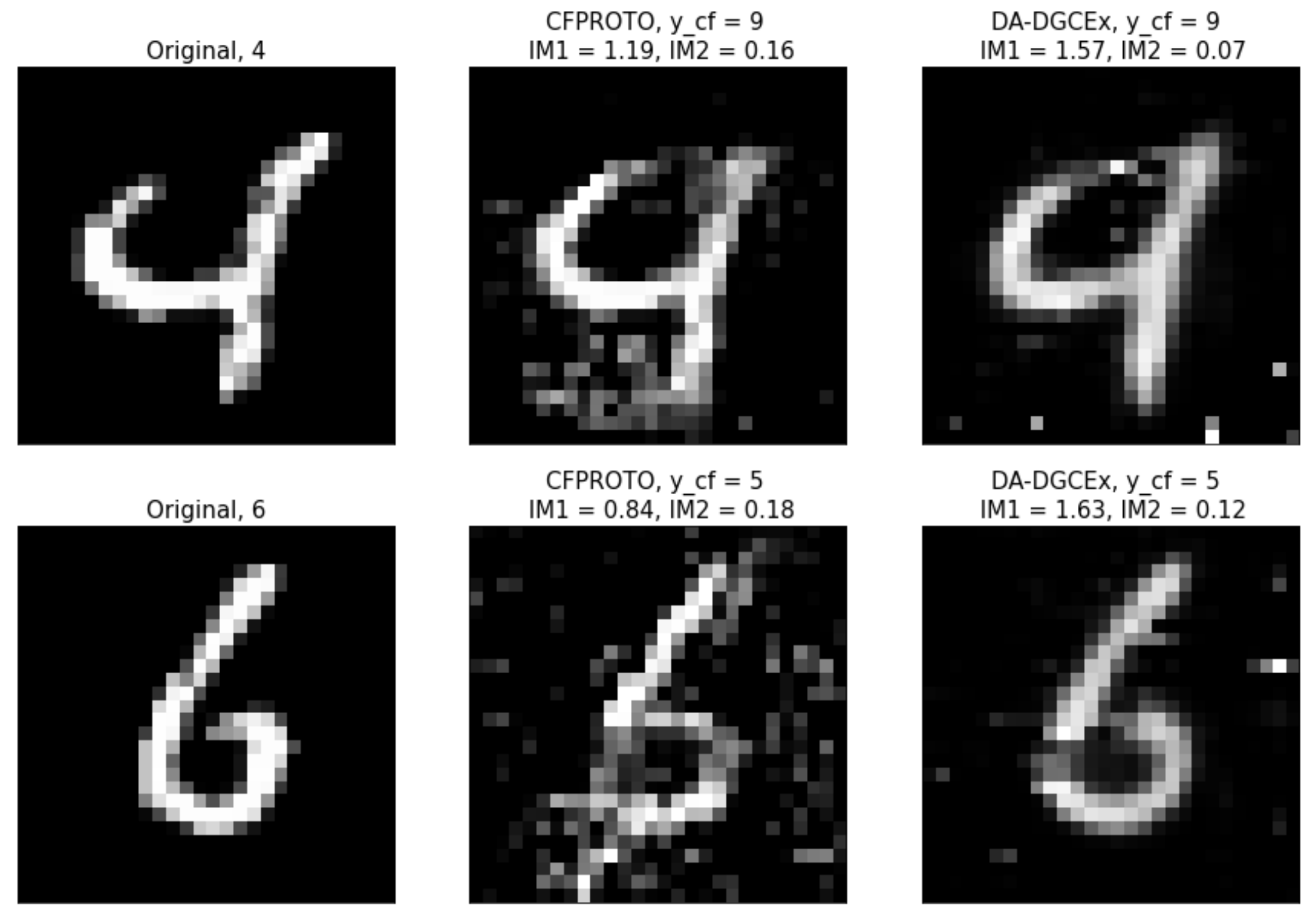}
    \captionsetup{width=.5\textwidth,skip=5pt}

    \caption{Examples of two counterfactual instances generated by CFPROTO and DA-DGCEx where being IM2 lower when using DA-DGCEx, IM1 is higher.}
    \label{fig:im1_fail}
\end{figure}

Analyzing the metrics used, we have seen that IM1 does not measure the interpretability of counterfactual instances well in some cases, and often a lower value of IM1 does not indicate that the counterfactual instances are more interpretable. In the figure, we have visualized two examples of counterfactual instances generated with CFPROTO and DA-DGCEx. In these examples, it can be seen that the instances generated with CFPROTO are less realistic than those generated with DA-DGCEx. This is reflected in the value of IM2, since those of CFPROTO have a higher value of IM2 than those of DA-DGCEx. However, it can be seen that even if the instances generated by CFPROTO are not realistic, the value in terms of IM1 is much lower than that of DA-DGCEx, which does not reflect the reality.  


\subsection{Experimental Framework}
The experiments are run on an Nvidia-Docker container that uses ubuntu 18.04. The models were implemented using the Keras library of Tensorflow. The Tensorflow version used in this case is Tensorflow 2.3.0-rc0. The optimization of the models was performed using Adam. For training the models an NVIDIA TITAN V GPU has been used, with a memory of 12 GB, in an Intel i7-6850K 3.6Ghz machine with 32 GB of DDR4 RAM.
\section{Conclusions and Future Work} \label{sec:conclusions}

In this work, Distribution Aware Deep Guided Counterfactual Explanations (DA-DGCEx) is proposed. This method is a variation of DGCEx, which improves the counterfactual generation process by adding a term in the loss function that penalizes counterfactual instances falling outside the data distribution. In this way, the proposed method provides in-distribution straightforward counterfactual explanations by combining AEs with deep learning classifiers. In addition, the generated counterfactual explanations are user-driven, since DA-DGCEx allows indicating the class to which the input is going to be changed.

DA-DGCEx has been validated in the MNIST handwritten digit classification dataset, and the results obtained have been compared with CFPROTO and DGCEx. The IM1 and IM2 metrics were used for the evaluation. As for IM1, we have seen that both CFPROTO and DA-DGCEx outperform DGCEx. However, between CFPROTO and DA-DGCEx we have not seen many differences, since in half of the cases one is better than the other and vice versa. Despite this, we have seen that in practice IM1 does not provide much information regarding the interpretability of the generated counterfactual instances, since in some cases where the IM1 was low, the generated instances were not realistic. On the other hand, as for IM2, we have seen that DA-DGCEx outperforms the other two methods, having an improvement of 0.29 over CFPROTO and 0.21 over DGCEx, while reducing variability. Furthermore, we have seen that the probability of obtaining better counterfactual instances in terms of IM2 using DA-DGCEx over CFPROTO and DGCEx is 0.62 and 0.76 respectively. Moreover, the speed at which DGCEx and DA-DGCEx generate the counterfactual instances is much higher than CFPROTO, as the first two take 0.14 s/it on average and CFPROTO takes 17.4 s/it.


In conclusion, DA-DGCEx is able to generate user-driven counterfactual explanations close to the data manifold, without the need to solve an optimisation problem for each input to be explained.

As future work, we plan to apply DA-DGCEx for anomaly or defect detection industrial scenarios. That is, on the one hand we want to know if DA-DGCEx is able to generate counterfactual explanations that help us to detect different types of defects in image-data industrial scenarios. And on the other hand, we want to do the same in time-series data, to understand the reason of different anomalies.
\bibliographystyle{IEEEtran}

\bibliography{references}

\end{document}